\documentclass[journal]{IEEEtran}

\usepackage{graphicx}
\usepackage{caption}
\usepackage{multirow}
\usepackage{mathtools}
\usepackage{amsmath}
\usepackage{algorithm}
\usepackage{algorithmic}
\usepackage{amssymb}
\usepackage{xcolor}
\usepackage{tikz}
\usepackage{pgfplots}
\pgfplotsset{compat=1.7}
\usepackage{tikz-3dplot}
\ifCLASSINFOpdf
\else
\fi
\begin{document}
%
\title{An Attention-based ConvLSTM Autoencoder with Dynamic Thresholding for Unsupervised Anomaly Detection in Multivariate Time Series}
%
%
%

\author{Tareq Tayeh\IEEEauthorrefmark{1},
        Sulaiman Aburakhia\IEEEauthorrefmark{1},
        Ryan Myers\IEEEauthorrefmark{2},
        and Abdallah Shami\IEEEauthorrefmark{1}\\
        ECE Department, Western University, London, Canada \IEEEauthorrefmark{1}, National Research Council Canada, London, Canada \IEEEauthorrefmark{2}\\
\{ttayeh, saburakh, abdallah.shami\}@uwo.ca,\\
ryan.myers@nrc-cnrc.gc.ca}

\maketitle

\begin{abstract}
As a substantial amount of multivariate time series data is being produced by the complex systems in Smart Manufacturing (SM), improved anomaly detection frameworks are needed to reduce the operational risks and the monitoring burden placed on the system operators. However, building such frameworks is challenging, as a sufficiently large amount of defective training data is often not available and frameworks are required to capture both the temporal and contextual dependencies across different time steps while being robust to noise. In this paper, we propose an unsupervised Attention-based Convolutional Long Short-Term Memory (ConvLSTM) Autoencoder with Dynamic Thresholding (ACLAE-DT) framework for anomaly detection and diagnosis in multivariate time series. The framework starts by pre-processing and enriching the data, before constructing feature images to characterize the system statuses across different time steps by capturing the inter-correlations between pairs of time series. Afterwards, the constructed feature images are fed into an attention-based ConvLSTM autoencoder, which aims to encode the constructed feature images and capture the temporal behavior, followed by decoding the compressed knowledge representation to reconstruct the feature images input. The reconstruction errors are then computed and subjected to a statistical-based, dynamic thresholding mechanism to detect and diagnose the anomalies. Evaluation results conducted on real-life manufacturing data demonstrate the performance strengths of the proposed approach over state-of-the-art methods under different experimental settings.
\end{abstract}


%
\IEEEpeerreviewmaketitle

\section{Introduction}
%
%
%
%

\IEEEPARstart{M}{anufacturing} has advanced massively in recent years and has become more computerized, complex, and automated, driving the emergence of Smart Manufacturing (SM) \cite{kusiak2018smart}. SM is a technology-driven approach that integrates Artificial Intelligence (AI), predictive analytics, big data, and Industrial Internet of Things (IIoT) to harness sensor data and automation for improving the manufacturing performance. As quality control is an indispensable part of the production process in all manufacturing industries around the globe, SM enables higher quality products to be manufactured, while reducing costs and improving safety \cite{kang2016smart}.

SM encompasses complex systems of interconnected sensors and computer components with diverse types that generate a substantial amount of multivariate time series data. As a result, potential system or production failures might occur, deeming anomaly detection at certain time periods as a vital task in order for the operators to solve the underlying issues. Anomalies within a sequence of data are broadly defined as observations that are unusual and signify irregular behavior. These irregular and unusual events have a very low probability of occurring, meaning that manual detection processes are a meticulous assignment that often requires more manpower than is generally available. Broken, cracked, and other imperfect products may result in costly returns, imposing operational and financial difficulties \cite{russo2017combined}. It is estimated that a 1\% productivity improvement across the manufacturing industry can result in an annual savings of USD \$500 million, whereas a breakdown in the production line can cost up to USD \$50 thousand per hour \cite{ProgressDataRPM}. Furthermore, predicting anomalies on time can decrease the number of breakdowns by up to 70\%,  maintenance costs by up to 30\%, and over scheduled repairs by up to 12\% \cite{ProgressDataRPM}. As a result, Machine Learning (ML) has been utilized in recent years to detect anomalies \cite{yu2017recursive, injadat2020multi, yang2019tree}, and in particular, Deep Learning (DL) algorithms \cite{kieu2018outlier, aburakhia2020transfer, guo2018multidimensional}.

Automated anomaly detection algorithms leverage DL due to the latter's data-driven nature, efficiency, and ability to perform data analysis without explicitly programming the application \cite{Goodfellow-et-al-2016}. Moreover, DL is able to learn higher level features from data in a hierarchical fashion while continuously improving the system's accuracy and automated decision making. The aforementioned features make DL one of the main contributors to the fast growth of SM, as it reduces operating costs and improves operations visibility \cite{wang2018deep}.

A basic requirement for building a supervised learning-based automated system to detect anomalies on a classification objective is the accessibility of a sufficient amount of training data for each class \cite{cook2019anomaly}. However, efficient supervised learning methods are often infeasible, as with well-optimized processes, there is often an abundance of non-anomalous data and a relatively small or no amount of anomalous data. To address this data imbalance challenge, a substantial amount of unsupervised anomaly detection methods have been developed in recent years, as these methods are trained on unlabeled input data with no output variables. An additional advantage of this approach is potentially detecting anomalies in novel classes that are not part of the training data set, providing a general solution to the surface quality manufacturing process task \cite{chalapathy2019deep}.


Some previous approaches in the literature use defective samples for training, not solving the anomaly detection task described in this work. Other approaches utilize classical methods, such as probabilistic, distance-based, clustering, ensemble, and predictive approaches to detect anomalies in an unsupervised fashion, but all fail to capture complex structures in the data without the input of domain experts \cite{chalapathy2019deep}. Unsupervised ML approaches were then proposed to capture these complex structures, however, they start failing to deal with the high dimensionality of the data feature space and the varying data aggregation as the data volume increases, requiring human expertise for feature extraction \cite{wang2018deep}. Various DL architectures, such as Convolutional Neural Networks (CNNs), Long Short-Term Memory (LSTM) neural networks, Convolutional LSTM (ConvLSTM) \cite{shi2015convolutional} neural networks and autoencoders have emerged in recent literature to solve the aforementioned challenges. Furthermore, when a production failure is occurring, it is important to localize the anomaly root causes to plan adequate countermeasures and fix the production system. This is done by pinpointing the sensors with anomalous readings, assisting the system operators to perform the system diagnosis and repair the system accordingly.

In this paper, we propose the ACLAE-DT framework, an unsupervised attention-based ConvLSTM autoencoder with dynamic thresholding to detect anomalies and identify anomalies' root cause in a manufacturing process multivariate timeseries. ACLAE-DT is a DL-based framework that is able to capture both the temporal and contextual dependencies across different time steps in the multivariate time series data, while being robust to noise. The framework starts by normalizing the input data via Min-Max scaling to scale the values to a fixed range. Post pre-processing, the data is then enriched using sliding windows to be able to capture more temporal behavior via the lagged values, followed by the incorporation of contextual manufacturing process data via an embedding layer to capture the contextual information. Afterwards, ACLAE-DT constructs feature images based on the processed and enriched data, which are matrices of inner-products between a pair of time series within the sliding window segments. Feature images characterizes the system statuses across different time steps by capturing shape similarities and inter-correlations between two time series across different time steps, while being robust to noise. Subsequently, an attention-based ConvLSTM autoencoder is employed to encode the constructed feature images and capture the temporal behavior, followed by decoding the compressed knowledge representation to reconstruct the feature images input. The structures that exist in the data are learned and consequently leveraged when forcing the input through the autoencoder's bottleneck. Attention is added to the autoencoder to sustain a constant performance as the input time-series sequences increase, reducing model errors \cite{bahdanau2014neural}. Furthermore, several hyperparameters are optimized via random search in order to enhance the model's performance \cite{yang2020hyperparameter}. A dynamic thresholding mechanism is then utilized against the computed reconstruction errors to detect and diagnose the anomalies, where the threshold is dynamically updated based on statistical derivations from the reconstructed normal data errors. The intuition is that ACLAE-DT will not be able to reconstruct the feature images well if it never observed similar system statuses before, resulting in anomalous processes to be flagged as it crosses the computed threshold. ACLAE-DT underwent rigorous empirical analysis, where results demonstrated the superior performance of the proposed approach over state-of-the-art methods.

The work presented in this paper is able to capture both the temporal and contextual dependencies across different time steps in the multivariate time series data in a manufacturing process to detect anomalies and identify the anomalies' root cause. The main contributions of this paper include:
\begin{itemize}
  \item A novel framework that consists of pre-processing and enriching the multivariate time series, constructing feature images, an attention-based ConvLSTM network autoencoder to reconstruct the feature images input, and a dynamic thresholding mechanism to detect anomalies and identify anomalies' root cause in multivariate time series.
  \item A generic, unsupervised learning framework that utilizes state-of-the-art DL algorithms and can be applied for various different multivariate time series use cases in SM. 
  \item An attention-based, time-distributed ConvLSTM encoder-decoder model that is capable of sustaining a constant performance as the rate of input time-series sequences from the manufacturing operations increase. 
  \item A nonparametric and dynamic thresholding mechanism for evaluating reconstruction errors that addresses non-stationarity, noise and diversity issues in the data. 
  \item A robust framework evaluated on a real-life public manufacturing data set, where results demonstrated its performance strengths over state-of-the-art methods under various experimental settings.
\end{itemize}


The remainder of this paper is structured as follows. Section \ref{relatedwork} presents the motivations behind the use of DL in SM and explores related work. Section \ref{preliminaries} discusses the technical preliminaries of the key concepts used in this paper. Section \ref{framework} details the methodology and implementation of the ACLAE-DT framework. Section \ref{experiments} describes the data set used, the different experimental setups, and the comparison benchmarks. Section \ref{results} discusses the obtained results and performance evaluation. Finally, Section \ref{conclusion} concludes the paper and discusses opportunities for future work.


\section{Motivation and Related Work}\label{relatedwork}

Robotic finishing is one of the most important applications in SM. The goal of robotic finishing is to manufacture products without any defects and with the adequate amount of surface roughness, in order to ensure smooth functional and financial operations. However, potential system or production failures might occur at any point in time, demonstrating the importance of having an efficient anomaly detection algorithm in place overseeing different parts of the system to detect and mitigate any manufacturing malfunctions.

Let us consider a robotic finishing system designed and calibrated to produce a finished metal car part. Multiple sensors will be mounted on the system at different locations in order to gather different types of readings, such as the feed rate, spindle rate, and torque. As the robotic machine is manufacturing the car part, all the sensor readings are being processed in order to monitor the system's performance. In the case of an imperfect finished car part or an anomalous system behavior, the anomaly detection algorithm will trigger an alarm and would ideally locate the anomalous sensor readings or system behavior, in an effort to notify the system operators and enable them to solve the underlying issues.

Many classical methods have been used to detect anomalies, such as probabilistic approaches \cite{saci2021autocorrelation}, distance-based approaches \cite{huo2019anomalydetect}, clustering approaches \cite{LI2021106919}, and predictive approaches \cite{gokcesu2017online}. These approaches may be computationally incomplex, however, their performance is sub-optimal as they fail to capture complex structures in the data without the input of domain experts \cite{chalapathy2019deep}. Furthermore, as the data volume increases, traditional approaches can fail to scale up as required to maintain their anomaly detection performances \cite{chalapathy2019deep}. 

Moreover, ML algorithms were proposed to resolve the limitations in classical methods for anomaly detection. ML algorithms include K-Nearest Neighbors (K-NNs) \cite{xie2012scalable}, Support Vector Machines (SVMs) \cite{wang2004anomaly}, and neural networks \cite{ryan1998intrusion}. These algorithms are all supervised learning-based, meaning they rely on labeled normal and anomalous historical data for training \cite{rasheed2019anomaly}. However, collecting labeled anomalous data is often infeasible, as with well-optimized processes, there is a often a high imbalance in the training data due to the abundance of non-anomalous samples and a relatively small or no amount of anomalous samples. Furthermore, ML approaches often require human input for feature extraction, where they start to fail in dealing with the dimensionality and variety of the data as data volume and velocity increases \cite{wang2018deep}. 

DL techniques have emerged in recent literature to solve the aforementioned challenges. Although ML is a data-driven AI technique to model input and output relationships, DL has distinctive advantages over ML in terms of feature learning, model construction, and model training. DL transforms data into abstract representations, allowing hierarchical discriminative features to be learned, which eliminates any manual feature development by domain experts. Moreover, DL integrates model construction and feature learning into a single model and trains the model's parameters jointly, creating an end-to-end learning structure with minimal human interference.

There has been an increase in available types of sensory data collected from various distinct aspects across the operational system in SM. As a result, data modelling and analysis are vital tasks in order to handle the large data volume increase and the real-time data processing to detect any system anomalies; tasks which DL excel in \cite{kusiak2017smart}. Different DL architectures were used in the literature to detect anomalies in SM, such as CNNs \cite{tayeh2020distance}, LSTM neural networks \cite{feng2017multi}, and autoencoders \cite{bayram2021real}. A CNN uses convolution in place of general matrix multiplication in at least one of their layers, and reduces the number of parameters very efficiently without losing out on the quality of models. This makes a CNN the prime choice for analyzing visual imagery, however, it faces difficulty in learning the high-dimensional features of the input time series as it operates in vector space. However, LSTMs, a special kind of Recurrent Neural Networks (RNNs), excel in modelling temporal behavior, as they use a feedback loop for learning and an extra parameter metric for connections between time-steps. Leveraging the complementary strengths of CNNs and LSTMs, ConvLSTMs \cite{shi2015convolutional} emerged to accurately model the spatio-temporal information by having convolutional structures in both the input-to-state and state-to-state transitions.

\section{Technical Preliminaries}\label{preliminaries}

\subsection{Convolutional Neural Network (CNN)}
A CNN is a feedforward deep neural network that is most commonly applied to analyzing visual imagery \cite{valueva2020application}, having a wide range of applications such as image and video recognition, image classification, image segmentation, and recommender systems. A CNN uses convolution in place of general matrix multiplication in at least one of their layers, and reduces the number of parameters very efficiently without losing out on the quality of models. 



The network keeps on learning new higher dimensionality and more complex features with every layer. The input data layer takes an input vector with shape ($N \times H \times V \times D$), where $N$ is the number of images, $H$ is the height, $V$ is the width, $D$ is the number of channels. The input is then passed to the convolutional layer, which convolves the input and abstracts it to a feature map based on a set of weights called the kernel. In the convolutional layer, each neuron receives input from a restricted area of the previous layer called the neuron's receptive field. The receptive field is typically a square matrix of weights with sizes that are smaller than the input. The convolution operation conducted by the receptive field along the input area is described as:

\begin{equation}
    y_{ij} = \sigma \bigg( \sum_{r=1}^{F} \sum_{k=1}^{F} w_{rk}x_{(r+1 \times S)(k+j \times S)} + b \bigg)
    \label{convolutionoperation}
\end{equation}

\begin{equation}
    0 \leq i \leq \frac{H - F}{S}, 0 \leq j \leq \frac{V - F}{S}
    \label{convolutioncondition}
\end{equation}

where $y_{ij}$ is the feature map output of the node, $\sigma$ is the nonlinear activation function applied, $F$ is the height and width size of the receptive field, $r$ and $k$ are the receptive field's width and height step respectively, $w$ is the weight, $x$ is the input data, $S$ is the stride length, and $b$ is the bias vector.

The convolutional layer applies a filter that scans the image a couple of pixels at a time, reducing the single input image to produce a feature map of size $\big( (\frac{H-F+2P}{S+1}) \times (\frac{V-F+2P}{S+1}) \times K \big)$, where K indicates the number of filters and P indicates the padding value. The pooling layer scales down the information produced from the convolution layer and controls overfitting, by reducing the spatial dimension while maintaining the most essential information. Different pooling types include max pooling, where the maximum pixel value of a batch is selected, min pooling, where the minimum pixel value of a batch is selected, and average pooling, where the average value of the pixels in a batch is selected. The fully connected layer connects every neuron in one layer to every neuron in another layer.


\subsection{Long  Short-Term  Memory (LSTM) Neural Network}
\begin{figure}[htbp]
\centerline{\includegraphics[width=9cm]{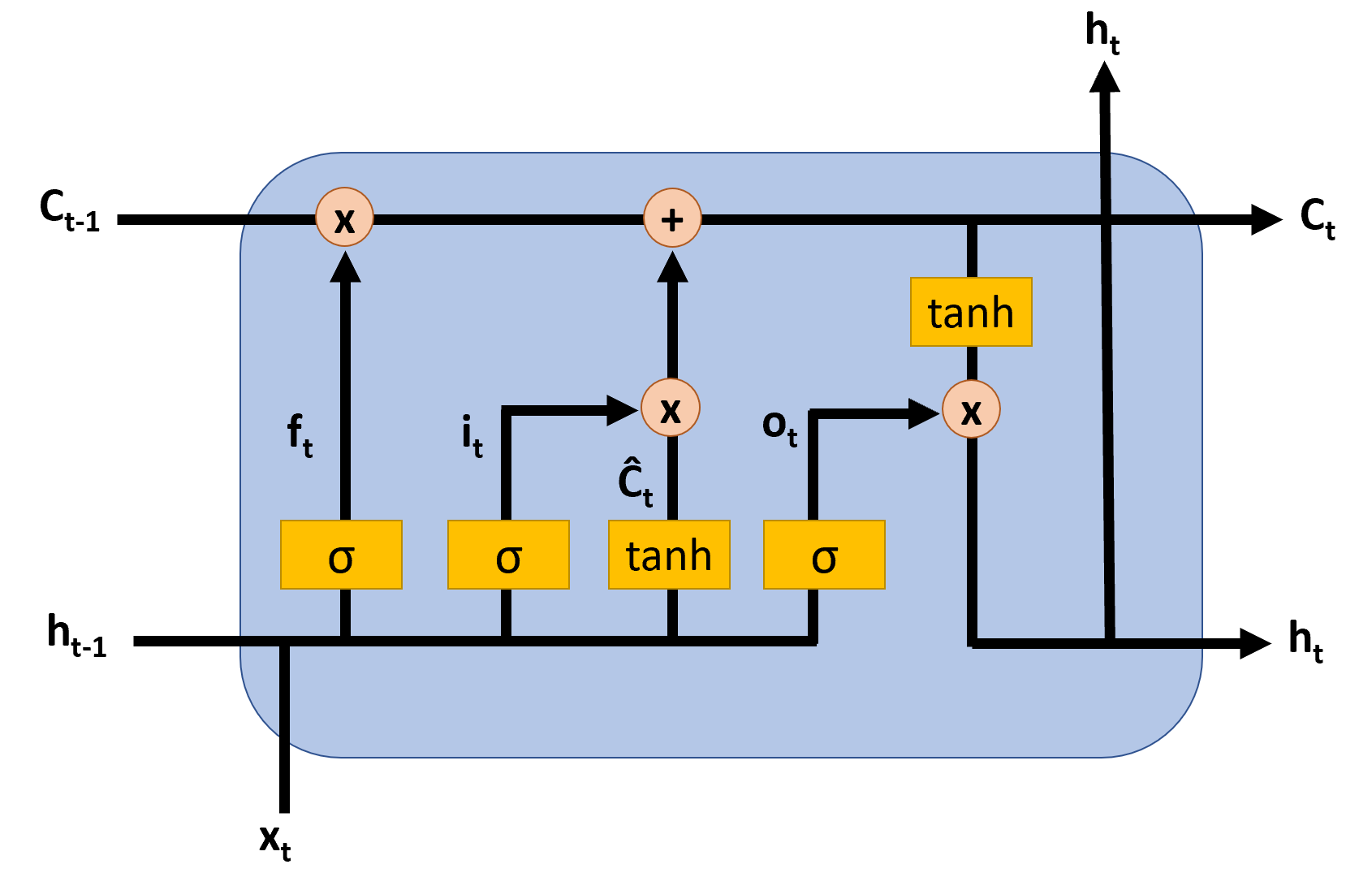}}
\caption{Structure of an LSTM memory cell}
\label{fig:LSTM}
\end{figure}

An LSTM deep neural network is a special variant of RNN that excel in modelling temporal behavior, such as time-series, language, audio, and text, due to the feedback loops used for learning and the extra parameter metric available for connections between time-steps. An LSTM’s main components are the memory cells and the input, forget, and output gates. These components allow the LSTM network to have connections from previous time steps and layers, where every output is influenced by the input as well as the historical inputs. Usually, there are multiple LSTM layers, where each layer is comprised of many LSTM units, and each unit comprise of input, forget, and output gates. The input gate protects the unit from irrelevant events, the forget gate helps the unit forget previous memory contents, and the output gates exposes the contents of the memory cell at the output of the LSTM unit. The equations for the input, forget, and output gates, as well as the candidate cell state, the cell state, and the LSTM cell output are described as, respectively:







\begin{gather}
    i_{t} = \sigma \big( W_{hi}h_{t-1} + W_{xi}x_{t} + b_{i} \big)
    \label{inputgate} \\
    f_{t} = \sigma \big( W_{hf}h_{t-1} + W_{xf}x_{t} + b_{f} \big)
    \label{forgetgate} \\
    o_{t} = \sigma \big( W_{ho}h_{t-1} + W_{xo}x_{t} + b_{o} \big)
    \label{outputgate} \\
    \tilde{C}_{t} = tanh \big( W_{hC}h_{t-1} + W_{xC}x_{t} + b_{C} \big)
    \label{candidatecellstate} \\
    C_{t} = f_{t} C_{t-1} + (1-f_{t}) \tilde{C}_{t}
    \label{cellstate} \\
    h_{t} = o_{t} tanh\big(C_{t}\big)
    \label{celloutput} 
\end{gather}

where $i$ is the input gate, $f$ is the forget gate, $o$ is the output gate, $\tilde{C}_{t}$ is the candidate cell state, $C$ is the cell state, $h$ is the hidden state and cell output, $\sigma$ denotes a logistic sigmoid function, $W$ is the weight matrix, and $b$ is the bias vector. Figure \ref{fig:LSTM} visualizes the structure of an LSTM memory cell.

\subsection{Convolutional LSTM (ConvLSTM)}

\begin{figure}[htbp]
\centerline{\includegraphics[width=9cm]{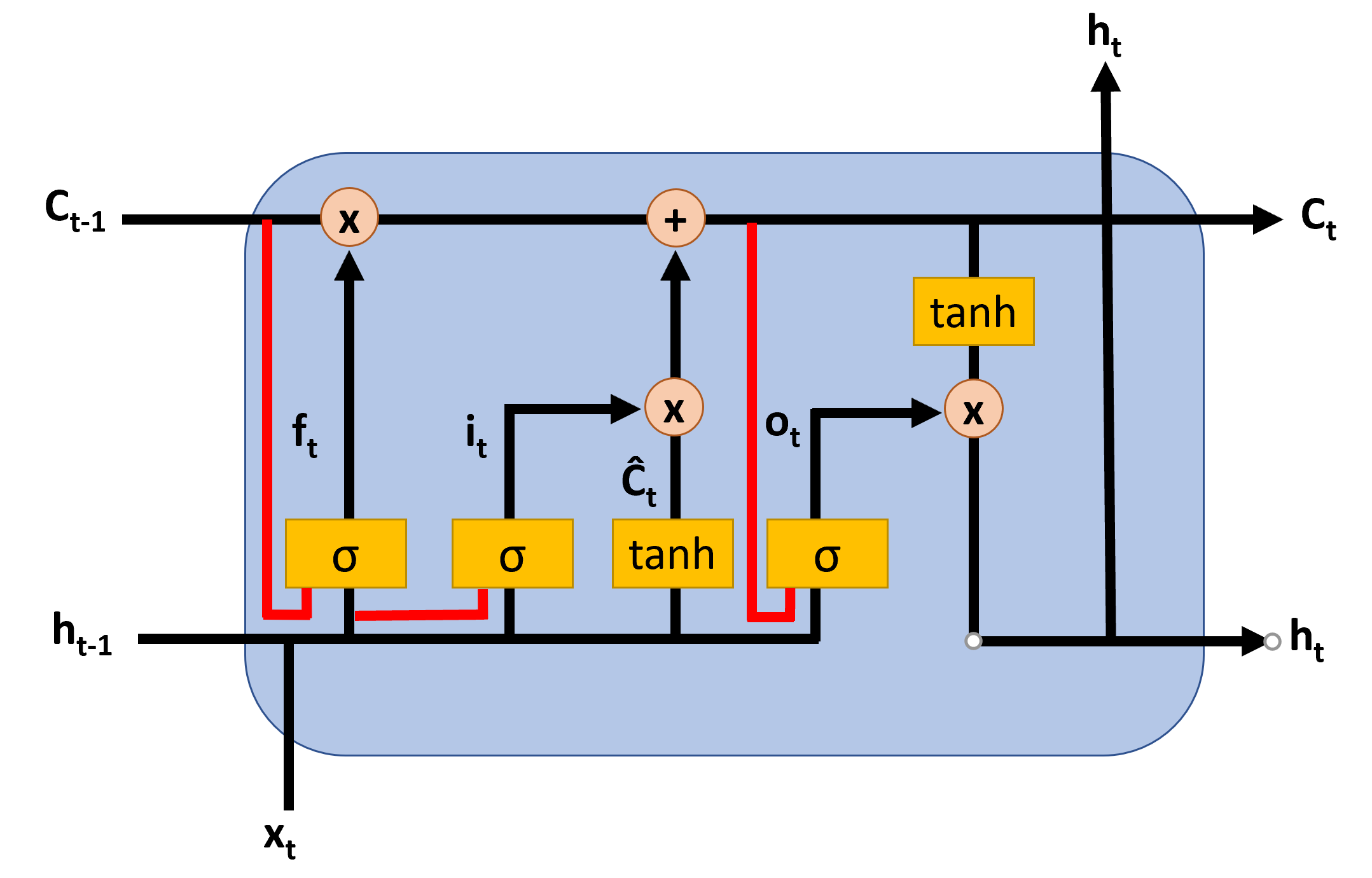}}
\caption{Structure of a ConvLSTM memory cell}
\label{fig:ConvLSTM}
\end{figure}

ConvLSTMs \cite{shi2015convolutional} are a special kind of LSTMs that emerged to accurately model the spatio-temporal information, by leveraging the strengths of a CNN and an LSTM. Similar to the LSTM, the ConvLSTM is able to decide what information to discard or retain from the previous cell state in its present cell state. However, convolutional structures are utilized in both the input-to-state and state-to-state transitions, which basically exchanges the internal matrix multiplications with convolution operations. The input vector to the ConvLSTM is fed as a series of 2-D or 3-D images, as the convolution operations allows the data that flows through the ConvLSTM cells to keep their input dimension instead of being a 1D vector with features. Furthermore, the transition between the states is analogous to the movement between the frames in an ConvLSTM. To describe the ConvLSTM operations, equations (3) - (8) are rewritten as:







\begin{gather}
    i_{t} = \sigma \big( W_{Ci} \circ C_{t-1} + W_{hi} \ast h_{t-1} + W_{xi} \ast x_{t} + b_{i} \big)
    \label{inputgateconvlstm} \\
    f_{t} = \sigma \big( W_{Cf} \circ C_{t-1} + W_{hf} \ast h_{t-1} + W_{xf}\ast x_{t} + b_{f} \big)
    \label{forgetgateconvlstm} \\
    o_{t} = \sigma \big( W_{Co}\circ C_{t} + W_{ho} \ast h_{t-1} + W_{xo} \ast x_{t} + b_{o} \big)
    \label{outputgateconvlstm} \\
    \tilde{C}_{t} = tanh \big( W_{hC} \ast h_{t-1} + W_{xC} \ast x_{t} + b_{C} \big)
    \label{candidatecellstateconvlstm} \\
    C_{t} = f_{t} \circ C_{t-1} + (1-f_{t}) \circ \tilde{C}_{t}
    \label{cellstateconvlstm} \\
    h_{t} = o_{t} \circ tanh\big(C_{t}\big)
    \label{celloutputconvlstm} 
\end{gather}

where $\circ$ represents the Hadamard product, $\ast$ represents the convolutional operator, $W_{Ci}$, $W_{hi}$, $W_{xi}$, $W_{Cf}$, $W_{hf}$, $W_{xf}$, $W_{Co}$, $W_{ho}$, $W_{xo}$, $W_{hC}$, $W_{xC}$ $\in \mathbb{R} ^{n \times T}$ represent the convolutional kernels within the model and $b_{i}, b_{f}, b_{o}, b_{C}$ are the bias parameters. Figure \ref{fig:ConvLSTM} visualizes the structure of a ConvLSTM, where the red lines indicate the extra connections found in a ConvLSTM cell over an LSTM cell.

\subsection{Autoencoder}
An autoencoder is an unsupervised feedforward neural network that imposes a bottleneck in the network, forcing a compressed knowledge representation of the original input. More specifically, an autoencoder learns how to efficiently compress and encode data, before learning how to reconstruct the data back from the reduced encoded representation to an output representation that is as close to the original input as possible. An autoencoder consists of 3 main parts: the encoder, which learns how to reduce the input dimensions and compress the input data into an encoded compressed representation, the compressed representation itself, and the decoder, which learns how to reconstruct the compressed representation to be as close to the original input as possible. The network is trained to minimize the reconstruction error, $L(x,\hat{x})$, which measures the differences between the original input and the consequent reconstruction. By design, autoencoders reduce data dimensionality by learning to ignore noise in the data.

\section{ACLAE-DT Framework}\label{framework}
\begin{figure}[htbp]
\centerline{\includegraphics[width=8cm]{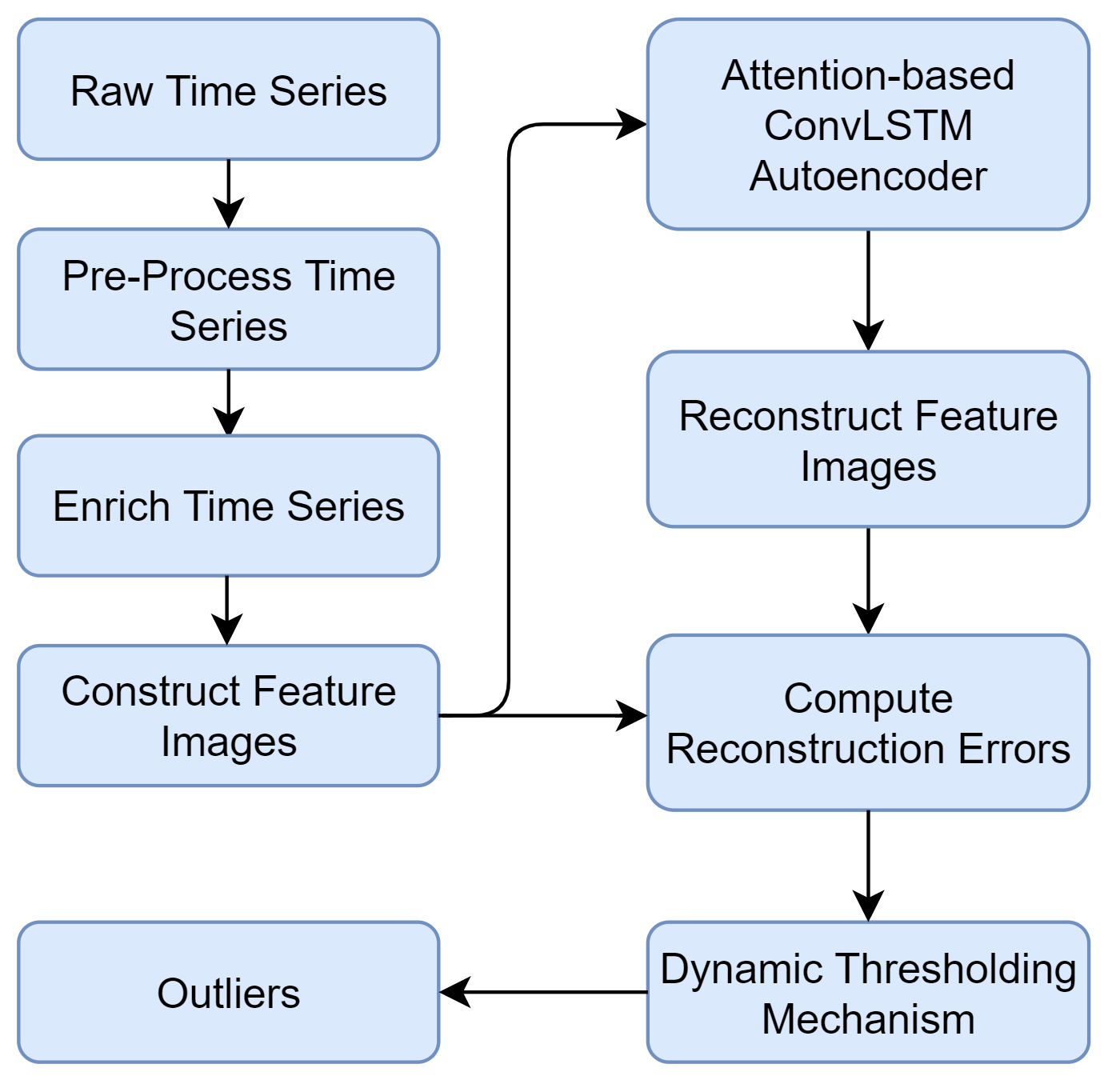}}
\caption{ACLAE-DT Framework}
\label{fig:Framework}
\end{figure}

The following section details the ACLAE-DT framework's design, methodology, and implementation. First of all, the problem statement addressed in this work is discussed, before detailing each module in ACLAE-DT. The framework starts off by pre-processing and enriching the input raw time series, before constructing feature images across the different time steps. Afterwards, the attention-based ConvLSTM autoencoder aims to reconstruct the feature images input by minimizing the reconstruction errors. Hyperparameter optimization is conducted to improve the model's performance. Lastly, a dynamic thresholding mechanism is applied against the computed reconstruction errors for anomaly detection and diagnosis. Figure \ref{fig:Framework} visualizes the ACLAE-DT framework.

\subsection{Problem Statement} 
The historical raw multivariate time series is represented as,

\begin{equation}
    X = (x_1, x_2, ... , x_n)^T \in \mathbb{R} ^{n \times T}
    \label{problemstatement}
\end{equation}

where $x_i$ is a single time series, $n$ is the number of time series, $T$ is the length of the time series, and $X$ is the entire time series. Assuming there are no anomalies in the training data, ACLAE-DT aims to detect anomalies by computing an anomaly score for each time step in $x_i$ after $T$, such that a score outside of the threshold boundaries is flagged as an anomaly, indicating an anomalous time step. Moreover, given the anomaly detection results, ACLAE-DT aims to identify the anomalies' root cause by quantitatively and qualitatively analyzing the time series that are most likely to be the causes of the flagged anomalous time step.

\subsection{Pre-Process Time Series}

Each input raw multivariate time series is normalized via Min-Max scaling to rescale their values between 0 and 1. Min-Max scaling normalization can be implemented as:

\begin{equation}
    x_i' = \frac{x_i - \text{min}(x_i)}{\text{max}(x_i) - \text{min}(x_i)} 
    \label{minmaxscaling}
\end{equation}

where $x'_i$ is the normalized input time series value. As each time series is considered a feature, feature scaling eliminates large scaled features to be dominating, while preserving all relationships in the data \cite{jayalakshmi2011statistical}. Furthermore, it allows gradient descent to converge much faster when training the attention-based ConvLSTM autoencoder model \cite{jayalakshmi2011statistical}.

\subsection{Enrich Time Series}
\subsubsection{Utilizing Sliding Windows}
Each pre-processed time series is then converted to a larger, more enriched time series of multivariate subsequences through the use of a sliding window. Sliding window indicates creating a specified window size and performing calculations on the data within this window, before sliding or rolling to the next window based on the step size specified and so on, till the entire data has been covered in at least a single window. This means data overlaps can occur across the different sliding windows when creating these extra sub periods and incorporating the lagged values, which can assist the attention-based ConvLSTM autoencoder to extract richer features from the constructed feature images \cite{yu2014time}. A more formal definition of a sliding window is specified as: given a time series $x_i$ of length $T$ and a user-defined subsequence length of $d$, all possible subsequences can be extracted by a sliding window of size $d$ across $x_i$ and considering each subsequence $s$ as $t-d$ to $t$, where $t$ indicates time position.


\subsubsection{Embedding Contextual Information}
Changes in a time series may be due to contextual changes. For example, a surface finished material in a manufacturing process could end up being scratched or bent due to the clamp pressure used to hold the workpiece in the vise rather than the x, y and z axis sensor values or the spindles. Therefore, taking contextual changes into consideration when detecting anomalies can enhance the anomaly detection performance \cite{gupta2013context}.

Contextual information are usually represented as text or categorical variables. However, DL-based models are only able to process numerical values. One method to achieve the required numerical conversion includes the use of ordinal encoding, where each unique category or text value is assigned an integer value. Nonetheless, this assumes that integer values have a natural ordered relationship between each other, which is often not the case. To resolve this shortcoming, one-hot encoding can be applied to remove the integer encoded variable and replace it with a new binary variable for each unique integer value. However, one-hot encoding does not scale well with respect to the number of categories, as the computation cost increase significantly as the categories increase, and does not capture the similarities between categories.

Entity embeddings \cite{guo2016entity} resolve the aforementioned limitations by using an additional mapping operation that transforms and represents each category to a low-dimensional space. The entity embedding vector or layer is a matrix of weights represented as $W_{embedding}$ $\in$ $\mathbb{R} ^{q \times p}$, where $q$ indicates the number of categories and $p$ indicates the number of dimensions in the low dimensional space. In this work, given a particular category $v$, a one-hot representation method $onehot(v)$ $\in$ $\mathbb{R} ^{q \times 1}$ is used. Afterwards, an embedded representation method $embedded(v) = $ $onehot(v)$ $\times$ $W_{embedding}$ is used for each category. It is important to set $p < q$ to ensure that as the number of categories increase, the dimensional value limits the computational cost increase. As a result, the embedded representation is much smaller than the one-hot representation and is able to capture similarities between the categories.

\subsection{Construct Feature Images}
\begin{figure}[htbp]
\centerline{\includegraphics[width=7cm]{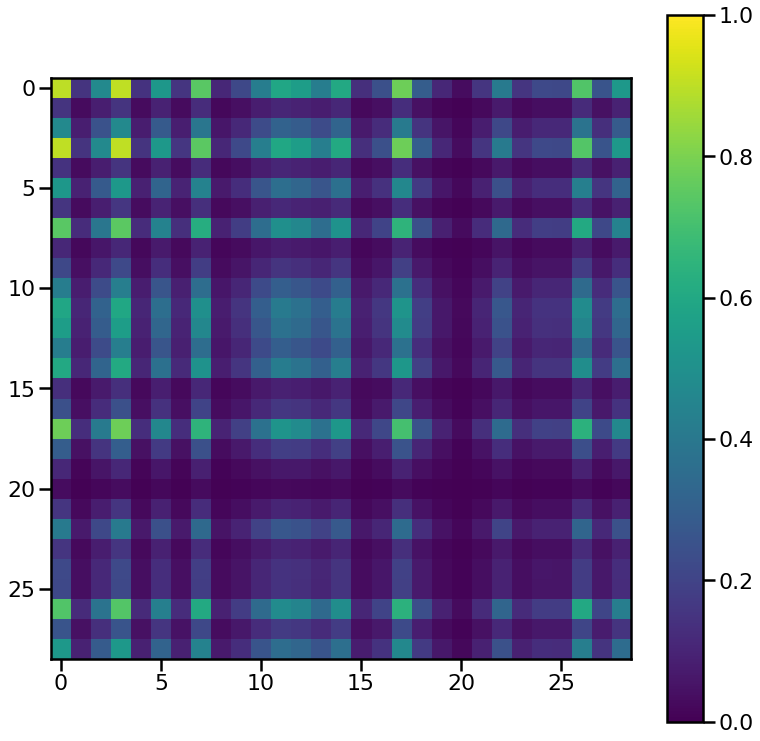}}
\caption{Feature Image Visualization}
\label{fig:feature_image}
\end{figure}

In order to characterize the manufacturing system status accurately, it is critical to calculate and pinpoint the correlations between the different pairs of time series \cite{hallac2017toeplitz}.
Acting as an extension to the work in \cite{zhang2019deep}, feature images are constructed in this paper to represent the inter-correlations between the different pairs of sensor values and contextual entity embeddings time series in a multivariate time series. More specifically, an ($n+p$) $\times$ ($n+p$) feature image $M^t$ is constructed for each sliding window segment $s$ based upon the pairwise inner-product of two time series within this segment. The inter-correlation between two time series in a single segment is calculated as:

\begin{equation}
    m^t_{ij} = \frac{ \sum_{\delta=0}^{w} x_i^{t-\delta} x_j^{t-\delta}}{s} \in M^t
    \label{featureimages}
\end{equation}

where $i$ and $j$ indicate two time series features, and $\delta$ indicates every single step in the segment. A feature images matrix is produced for each experiment or multivariate time series of length $T$, which consists of a feature image $M^t$ for each segment $s$. In addition to representing the inter-correlations and shape similarities between the pairs of time series, feature images are robust to input noise, as instabilities at certain time steps have small consequences. Figure \ref{fig:feature_image} visualizes a single feature image example $M^t$.


\subsection{Attention-based ConvLSTM Autoencoder Model}
\begin{figure*}
\centerline{\includegraphics[width=18cm]{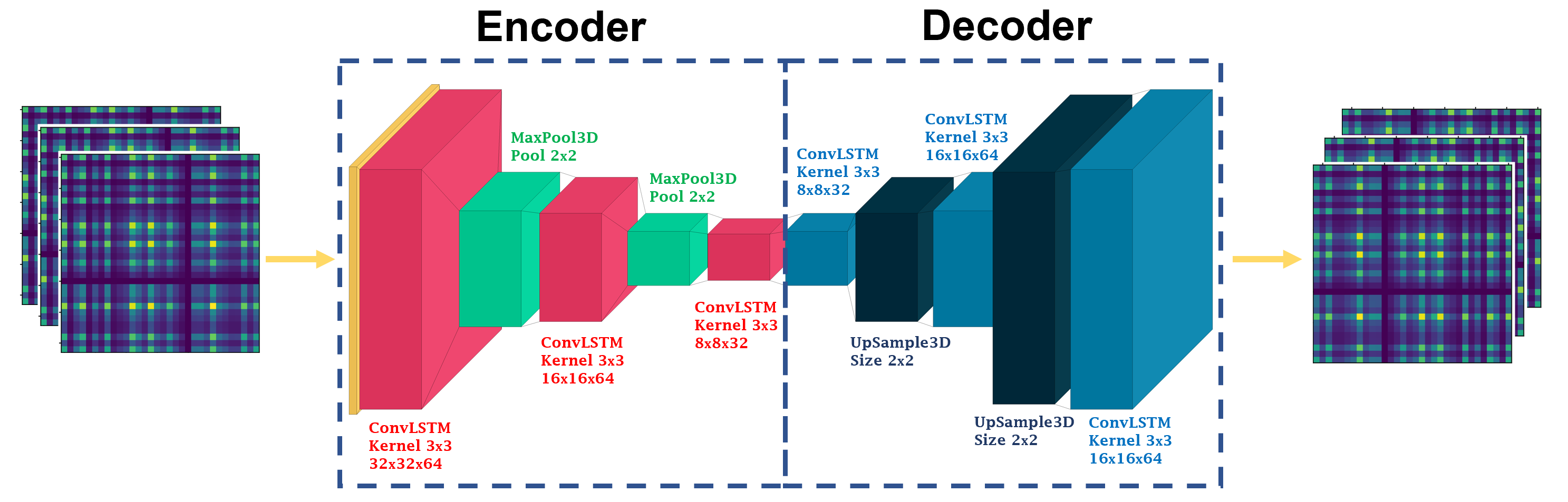}}
\caption{Proposed Attention-based ConvLSTM Autoencoder Model}
\label{fig:Model}
\end{figure*}

Once all feature images has been constructed, they are input into the ConvLSTM autoencoder for reconstruction. More specifically, the autoencoder first encodes the constructed feature images and captures the temporal behavior via three alternating ConvLSTM encoding layers and MaxPool layers, followed by decoding the compressed knowledge representation to reconstruct the original feature images input via three alternating ConvLSTM decoding layers and UpSample layers. All MaxPool and UpSample layers have a size of (2x2), whereas all the ConvLSTM layers have a kernel size of (3x3) and 'same' padding. Moreover, all the ConvLSTM layers have 64 filters, except for the last ConvLSTM encoder layer and the first ConvLSTM decoder layer, which have 32 filters. Figure \ref{fig:Model} visualizes the employed ConvLSTM autoencoder architecture.

Although ConvLSTMs have been developed to accurately model the spatio-temporal information in a sequence, its performance may deteriorate as the sequence length increases. To overcome this issue, the Bahdanau attention \cite{bahdanau2014neural} is added to the model, which can adaptively select relevant hidden states across different time steps and aggregate the representations of the informative feature maps to form a refined output of feature maps. A constant model performance is achieved and model errors are reduced as the input time-series sequences increase. The additive Bahdanau attention is described as:

\begin{gather}
    {c}^{t} = \sum_{t'=1}^{T_{x}} \alpha^{t,t'} h^{t'} \\
    \alpha^{t,t'} = \frac{exp(u^{t,t'})}{\sum_{k=1}^{T_{x}} exp(u^{t,k})} \\
    u^{t,t'} = a(s^{t-1},h^{t'})
\end{gather}

where $c$ is the context vector for the sequence of hidden state annotations, $\alpha$ denotes the weights of each annotation, and $u$ is the alignment model and output score of the feedforward neural network described by function $a$ that attempts to capture the alignment between the attention-based ConvLSTM hidden state $s$ of the previous time step $t-1$ and the $t'$-th annotation from the hidden state $h$ of the input sequence.



\subsection{Model's Hyperparameter Optimization}

In order to enhance the performance of the model, several hyperparameters need to be tuned and optimized \cite{yang2020hyperparameter}. Model hyperparameters are used in processes to help estimate the model parameters, which are learned and estimated during the training process of minimizing an objective loss function. Several hyperparameter optimization methods exist, such as grid search, random search, and gradient-based optimization. Grid search is an exhaustive search through a set of manually specified set of hyperparameter values, which is time-consuming and impacted by the curse of dimensionality \cite{claesen2014easy}. Gradient-based optimization utilizes the gradient descent algorithm to compute the gradient with respect to hyperparameters, but they only support continuous hyperparameters and can only detect a local optimum for non-convex hyperparameter optimization problems rather than a global optimum \cite{zoller2019benchmark}. Finally, random search randomly searches the grid space and supports all types of hyperparameters, allowing a larger and more diverse grid space to be explored. Hence, random search is used as the optimization method in this work as it is more computationally efficient than grid search and gradient-based optimization.

Five hyperparameters are optimized in this work, as presented in Table \ref{tab:hyperparameters} alongside all the values of consideration and in which layer they lie, if applicable. The selected hyperparameters are considered to be the most influential five hyperparameters on the model's performance, based on initial experiments. All hyperparameter value options are based off initial experiments as well. Activation function values considered include Rectified Linear Units (ReLU) \cite{nair2010rectified}, Leaky ReLU \cite{maas2013rectifier}, Exponential Linear Units (ELU) \cite{clevert2015fast}, and Scaled ELU (SELU) \cite{klambauer2017self}. Optimization algorithms considered include Adam \cite{kingma2014adam}, RMSProp \cite{hintonrmsprop}, AdaDelta \cite{zeiler2012adadelta}, and Stochastic Gradient Descent (SGD) \cite{kiefer1952stochastic}. Finally, the loss functions considered include Mean Absolute Error (MAE), Mean Squared Error (MSE), and Root MSE (RMSE).


\begin{table}[]
\begin{tabular}{|l|l|l|}
\hline
\textbf{Layer} & \textbf{Hyperparameter} & \textbf{Values}              \\ \hline
ConvLSTM       & Activation Function     & ReLU, Leaky RELU, ELU, SELU  \\ \hline
N/A            & Learning Rate           & 1e-2, 1e-3, 1e-4, 1e-5, 1e-6 \\ \hline
N/A            & Batch Size              & 16, 32, 64, 128, 256         \\ \hline
N/A            & Optimizer               & Adam, RMSProp, ADADelta, SGD \\ \hline
N/A            & Loss Function           & MAE, MSE, RMSE               \\ \hline
\end{tabular}
\caption{Model Hyperparameters}
\label{tab:hyperparameters}
\end{table}

\subsection{Compute Reconstruction Errors}
The model's loss objective in this work is to minimize the reconstruction errors of the normal feature images during training, in order to accurately reconstruct the normal feature images and inaccurately reconstruct the anomalous feature images during the testing phase. The loss function used is dependent on the random search hyperparameter optimization method output for the loss function hyperparameter.


\subsection{Dynamic Thresholding Mechanism}
At this stage, an efficient anomaly thresholding mechanism is needed in order to detect anomalous reconstructed feature images. Often, the anomaly threshold is learned with supervised methods, however, as the nature of the data in the manufacturing domain is continuously changing and there is insufficient labeled data for each class, supervised methods would not be optimal for use here \cite{chandola2009survey}. Hence, a nonparametric and dynamic anomaly thresholding mechanism is proposed in this work, which calculates a different threshold for each time series pair based on statistical derivations, achieving high anomaly detection performance with low overhead. More specifically, a single threshold is set against every single time series pair in the feature image, based on the mean and standard deviation of that specific normal time series pair reconstruction errors. Any time series pair value that surpasses the threshold at any time step during testing will be flagged as anomalous and will flag the entire process at that time step as anomalous. Mathematically, the method is described as:

\begin{equation}
    \epsilon_{ij} = \big(\mu(e_{ij}) + z\sigma(e_{ij})\big)^{T} \in \epsilon
\end{equation}

where $\epsilon_{ij}$ indicates the threshold value for the $i$ and $j$ time series features pair across the entire time series, $\epsilon$ is the ($n+p$) $\times$ ($n+p$) threshold matrix, $e_{ij}$ is the set of reconstruction errors for the normal $i$ and $j$ time series features pair, $\mu$ is the mean, $\sigma$ is the standard deviation, and $z$ is an ordered set of positive values representing the number of standard deviations. Values for $z$ depend on context, with a range of two to five found to produce the most accurate experimental results in this work. The presented dynamic thresholding mechanism detects outliers as well as localizes the anomaly root cause, by pinpointing the sensors that are causing the detected outlier.



\section{Experiments}\label{experiments}
The data set used in this paper is the Computer Numerical Control (CNC) Mill Tool Wear data set provided by the University of Michigan \cite{kovalenko2017smart} and found on Kaggle \cite{sun_2018}. The data set consisted of a series of machining experiments run on 2"x 2" x 1.5" wax blocks in a CNC milling machine in the System-level Manufacturing and Automation Research Testbed (SMART). Machining data was collected from a CNC machine for variations of feed rate, tool condition, and clamping pressure, where each experiment produced a finished wax part with an "S" shape. 44 time series readings from the 4 motors in the CNC machine, the X-axis, Y-axis, Z-axis, and spindle (S-axis), were collected at a sampling rate of 10 Hz. The data set contained a total of 25,286 time series measurements from the 18 experiments conducted, where 4 of these experiments failed the visual inspection check. 

In this paper, every measurement is taken as an independent observation within a sliding window to identify normal or anomalous behavior and to pinpoint the sensors that flagged windows as anomalous. Any measurement that is part of the 4 experiments with a failed visual inspection check contain at least a single anomalous time series reading.


In order to evaluate ACLAE-DT's anomaly detection performance, the attention-based ConvLSTM autoencoder model is compared with seven baseline methods. The baseline methods comprise of an ML classification method, a classical forecasting method, three state-of-the-art DL methods, and two variants of ACLAE-DT. The classical and ML methods are evaluated to demonstrate the effectiveness of using a DL model, the baseline DL methods are evaluated to demonstrate the effectiveness of a ConvLSTM autoencoder, and the two variants of ACLAE-DT are evaluated to demonstrate the effectiveness of each component within the model. The same number of layers, hyperparameters, and components are used for each DL method, if applicable. The baseline methods are:

\begin{enumerate}
\item SVM: An ML method that classifies whether a test data point is an anomaly or not based on the learned decision function from the training data.
\item Auto Regressive Integrated Moving Average (ARIMA): A classical prediction model that captures the temporal dependencies in the training data to forecast the predicted values of the testing data. 
\item LSTM Autoencoder: A DL method that utilizes LSTM networks in both the encoder and decoder.
\item ConvLSTM-LSTM Autoencoder: A DL method that utilizes ConvLSTM networks in the encoder and LSTM networks in the decoder.
\item CNN-LSTM Autoencoder: A DL method that utilizes CNN-LSTM networks in both the encoder and decoder.
\item ACLAE-DT Shallow: An ACLAE-DT variant that utilizes ACLAE-DT's model without the last MaxPool3D and ConvLSTM encoder components and without the first UpSample3D and ConvLSTM decoder components.
\item ACLAE-DT No-Attention: An ACLAE-DT variant that utilizes ACLAE-DT's model without attention.
\end{enumerate}

To empirically examine the models, three different experiments are conducted. The models are tested using a window size of 10 with a step size of 2 in Experiment 1, a window size of 30 with a step size of 5 in Experiment 2, and a window size of 60 with a step size of 10 in Experiment 3. All DL-based models are trained for 250 epochs. Moreover, comparison metrics are employed to evaluate the models used and compare their anomaly detection performances. In order to fully capture the values of the true and false positives and negatives for each model, the precision, recall, and F1 score metrics are utilized, as well as the time taken to train each model. An anomalous window is defined as a poorly reconstructed feature image with a value that surpasses the corresponding threshold. True positives in this work indicate anomalous windows correctly classified as anomalous and true negatives indicate non-anomalous windows correctly classified as non-anomalous. All experiments are repeated five times and the average results are computed for performance comparison. 

Afterwards, ACLAE-DT's results are analyzed to pinpoint the readings that flagged windows as anomalous. It is important to localize the anomaly root cause during a production failure to plan adequate countermeasures and fix the system.

All networks are built and implemented in Python 3.7.9, using the Tensorflow \cite{abadi2016tensorflow} and Keras \cite{chollet2015keras} libraries. All work is run on a machine which comprises of an NVIDIA GeForce GTX 1650 4GB, a 16GB DDR4 2666MHz RAM, and a 9th Generation Intel Core i7-9750H Processor.

\section{Performance Evaluation}\label{results}

\subsection{Anomaly Detection Results}
The anomaly detection performance for each model under the three different experimental settings are illustrated in Tables \ref{tab:evaluation10_2}, \ref{tab:evaluation30_5} and \ref{tab:evaluation60_10}, respectively. Note that the results for ARIMA and SVM are the same across all three experiments, as they do not consider window sizes and step sizes in their algorithmic calculations. Table \ref{tab:evaluation10_2} demonstrates the performance evaluation of all eight models in Experiment 1. It can be observed that ARIMA detected anomalies better than SVM, indicating that the data set had a temporal dependency feature that cannot be captured by the classification method. However, all the DL-based methods performed better than ARIMA, indicating DL's strength in capturing more complex structures and model a finer multivariate temporal dependency and correlation from the data set. Furthermore, it can be observed that all variants of ACLAE-DT performed greater than the three baseline DL models based on every single evaluation metric used, while taking less time to train. The full ACLAE-DT model performed at least 4.8\% better in every single evaluation metric, while taking at least 22.9\% less time to train than the three baseline DL models. Moreover, the full ACLAE-DT model performed either similarly or better than the two variant baseline models, while taking 6.7\% less time to train than the shallow model but 3.1\% more time than the no-attention model.

\begin{table}[]
\centering
\scalebox{0.98}{
\begin{tabular}{|l|l|l|l|l|}
\hline
\textbf{Method}                                                      & \textbf{Precision} & \textbf{Recall} & \textbf{F1-Score} & \textbf{\begin{tabular}[c]{@{}l@{}}Train \\ Time (s)\end{tabular}} \\ \hline
\begin{tabular}[c]{@{}l@{}}SVM\\ (Linear Kernel)\end{tabular}        & 0.15               & 0.17            & 0.16              & 14                                                                 \\ \hline
\begin{tabular}[c]{@{}l@{}}ARIMA\\ (2,1,2)\end{tabular}              & 0.52               & 0.59            & 0.56              & 98                                                                 \\ \hline
\begin{tabular}[c]{@{}l@{}}LSTM \\ Autoencoder\end{tabular}          & 0.83               & 0.80            & 0.82              & 13,468                                                             \\ \hline
\begin{tabular}[c]{@{}l@{}}ConvLSTM-LSTM \\ Autoencoder\end{tabular} & 0.80               & 0.84            & 0.82              & 11,914                                                             \\ \hline
\begin{tabular}[c]{@{}l@{}}CNN-LSTM \\ Autoencoder\end{tabular}      & 0.84               & 0.84            & 0.84              & 10,136                                                             \\ \hline
\begin{tabular}[c]{@{}l@{}}ACLAE-DT \\ Shallow\end{tabular}          & 0.94               & 0.87            & 0.90              & 8,372                                                              \\ \hline
\begin{tabular}[c]{@{}l@{}}ACLAE-DT \\ No Attention\end{tabular}     & 0.95               & 0.87            & 0.91              & 7,574                                                              \\ \hline
\begin{tabular}[c]{@{}l@{}}ACLAE-DT\\ Full\end{tabular}              & 0.95               & 0.88            & 0.92              & 7,812                                                              \\ \hline
\end{tabular}}
\caption{Anomaly Detection Results Using a Window Size of 10 with a Step Size of 2}
\label{tab:evaluation10_2}
\end{table}

\begin{table}[]
\centering
\scalebox{0.98}{
\begin{tabular}{|l|l|l|l|l|}
\hline
\textbf{Method}                                                      & \textbf{Precision} & \textbf{Recall} & \textbf{F1-Score} & \textbf{\begin{tabular}[c]{@{}l@{}}Train \\ Time (s)\end{tabular}} \\ \hline
\begin{tabular}[c]{@{}l@{}}SVM\\ (Linear Kernel)\end{tabular}        & 0.15               & 0.17            & 0.16              & 14                                                                 \\ \hline
\begin{tabular}[c]{@{}l@{}}ARIMA\\ (2,1,2)\end{tabular}              & 0.52               & 0.59            & 0.56              & 98                                                                 \\ \hline
\begin{tabular}[c]{@{}l@{}}LSTM \\ Autoencoder\end{tabular}          & 0.82               & 0.83            & 0.83              & 15,932                                                             \\ \hline
\begin{tabular}[c]{@{}l@{}}ConvLSTM-LSTM \\ Autoencoder\end{tabular} & 0.79               & 0.84            & 0.81              & 8,274                                                              \\ \hline
\begin{tabular}[c]{@{}l@{}}CNN-LSTM \\ Autoencoder\end{tabular}      & 0.83               & 0.85            & 0.84              & 5,362                                                              \\ \hline
\begin{tabular}[c]{@{}l@{}}ACLAE-DT \\ Shallow\end{tabular}          & 0.91               & 0.89            & 0.90              & 3,388                                                              \\ \hline
\begin{tabular}[c]{@{}l@{}}ACLAE-DT \\ No Attention\end{tabular}     & 0.95               & 0.88            & 0.90              & 3,122                                                              \\ \hline
\begin{tabular}[c]{@{}l@{}}ACLAE-DT\\ Full\end{tabular}              & 0.96               & 0.90            & 0.93              & 3,234                                                              \\ \hline
\end{tabular}}
\caption{Anomaly Detection Results Using a Window Size of 30 with a Step Size of 5}
\label{tab:evaluation30_5}
\end{table}

\begin{table}[]
\centering
\scalebox{0.9}{
\begin{tabular}{|l|l|l|l|l|}
\hline
\textbf{Method}                                                      & \textbf{Precision} & \textbf{Recall} & \textbf{F1-Score} & \textbf{\begin{tabular}[c]{@{}l@{}}Train \\ Time (s)\end{tabular}} \\ \hline
\begin{tabular}[c]{@{}l@{}}SVM\\ (Linear Kernel)\end{tabular}        & 0.15               & 0.17            & 0.16              & 14                                                                 \\ \hline
\begin{tabular}[c]{@{}l@{}}ARIMA\\ (2,1,2)\end{tabular}              & 0.52               & 0.59            & 0.56              & 98                                                                 \\ \hline
\begin{tabular}[c]{@{}l@{}}LSTM \\ Autoencoder\end{tabular}          & 0.79               & 0.83            & 0.82              & 7,462                                                              \\ \hline
\begin{tabular}[c]{@{}l@{}}ConvLSTM-LSTM \\ Autoencoder\end{tabular} & 0.77               & 0.82            & 0.79              & 13,496                                                             \\ \hline
\begin{tabular}[c]{@{}l@{}}CNN-LSTM \\ Autoencoder\end{tabular}      & 0.84               & 0.85            & 0.85              & 5,152                                                              \\ \hline
\begin{tabular}[c]{@{}l@{}}ACLAE-DT \\ Shallow\end{tabular}          & 0.96               & 0.99            & 0.97              & 2,814                                                              \\ \hline
\begin{tabular}[c]{@{}l@{}}ACLAE-DT \\ No Attention\end{tabular}     & 0.97               & 0.99            & 0.98              & 1,638                                                              \\ \hline
\begin{tabular}[c]{@{}l@{}}ACLAE-DT\\ Full\end{tabular}              & 0.99               & 1.00            & 1.00              & 1,736                                                              \\ \hline
\end{tabular}}
\caption{Anomaly Detection Results Using a Window Size of 60 with a Step Size of 10}
\label{tab:evaluation60_10}
\end{table}

Similar observations can be drawn from Tables \ref{tab:evaluation30_5} and \ref{tab:evaluation60_10}, which demonstrate the performance evaluation of all eight models in Experiment 2 and Experiment 3, respectively. As the window size and step size increased, the training time for all ACLAE-DT variants decreased, whilst average performance improved. This was not the case with the DL baseline models however, as training time and performance did not follow a general trend as window sizes and step sizes changed. In Experiment 2, the full ACLAE-DT model performed at least 5.8\% better in every single evaluation metric, while taking at least 65.8\% less time to train than the three baseline DL models. In Experiment 3, the full ACLAE-DT model performed at least 16.5\% better in every single evaluation metric, while taking at least 196.7\% less time to train than the three baseline DL models. In both experiments, the full ACLAE-DT model performed better than the two variant baseline models, while taking less time to train than the shallow model but more time than the no-attention model. 

All the previous results demonstrate the strength of utilizing a DL-based anomaly detection model in multivariate time series, as SVM and ARIMA failed to capture complex relationships and detect anomalies appropriately. Moreover, the results demonstrate the effectiveness of deploying ConvLSTM networks in both the encoder and decoder in an autoencoder compared to deploying LSTM networks or CNNs in either the encoder or decoder, as ACLAE-DT was capable of capturing the inter-sensor correlations and temporal patterns of multivariate time series effectively. The results further demonstrate the effectiveness of constructing a deeper model and adding attention to it, particularly for when the window size is 30 and above, as performance constantly improved. The full ACLAE-DT model and all its variants performed the best in Experiment 3, whilst taking the least amount of training time. The full ACLAE-DT model had the best model performance out of all the compared models in the aforementioned experimental setting, scoring a perfect recall and F1-score, and close to perfect precision score, with a total average training time of 1,736 seconds.

\subsection{Anomaly Root Cause Identification Results}

\begin{figure}
\centerline{\includegraphics[width=9cm]{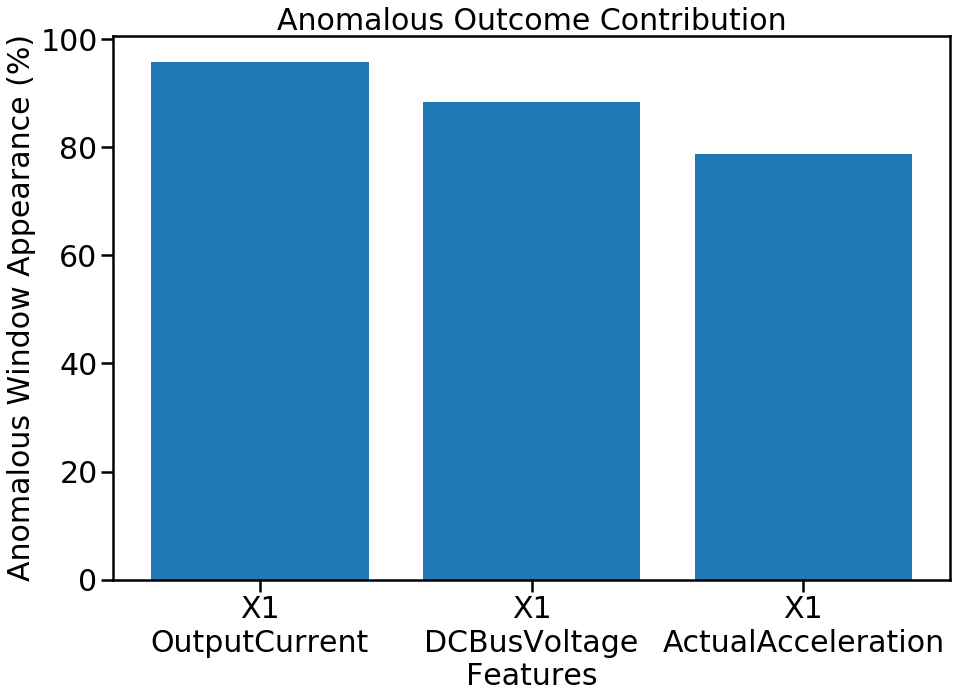}}
\caption{Anomaly Root Cause Feature Analysis}
\label{fig:anomaly_root_cause}
\end{figure}

If the reconstruction error of an inter-correlation between two time series crossed the set threshold for a particular window, then the corresponding pair of sensors were signified as contributors towards the anomalous window. The three sensor readings that contributed the most towards the flagged anomalous windows in Experiment 3 are visualized in Figure \ref{fig:anomaly_root_cause}. The x-axis indicates the sensor reading features, and the y-axis indicates the feature's anomalous window appearance percentage. It can be observed from the figure that the readings from the X-axis motor had the greatest influence on the success of the visual inspection check, as they contributed the most towards flagging a window as anomalous. The X1-OutputCurrent feature had the greatest influence as it passed its threshold in 95.7\% of the windows, followed by the X1-DCBusVoltage feature as it passed its threshold in 88.2\% of the windows, followed by the X1-ActualAcceleration feature as it passed its threshold in 78.7\% of the windows.



\begin{figure}
\centerline{\includegraphics[width=9cm, height=18cm]{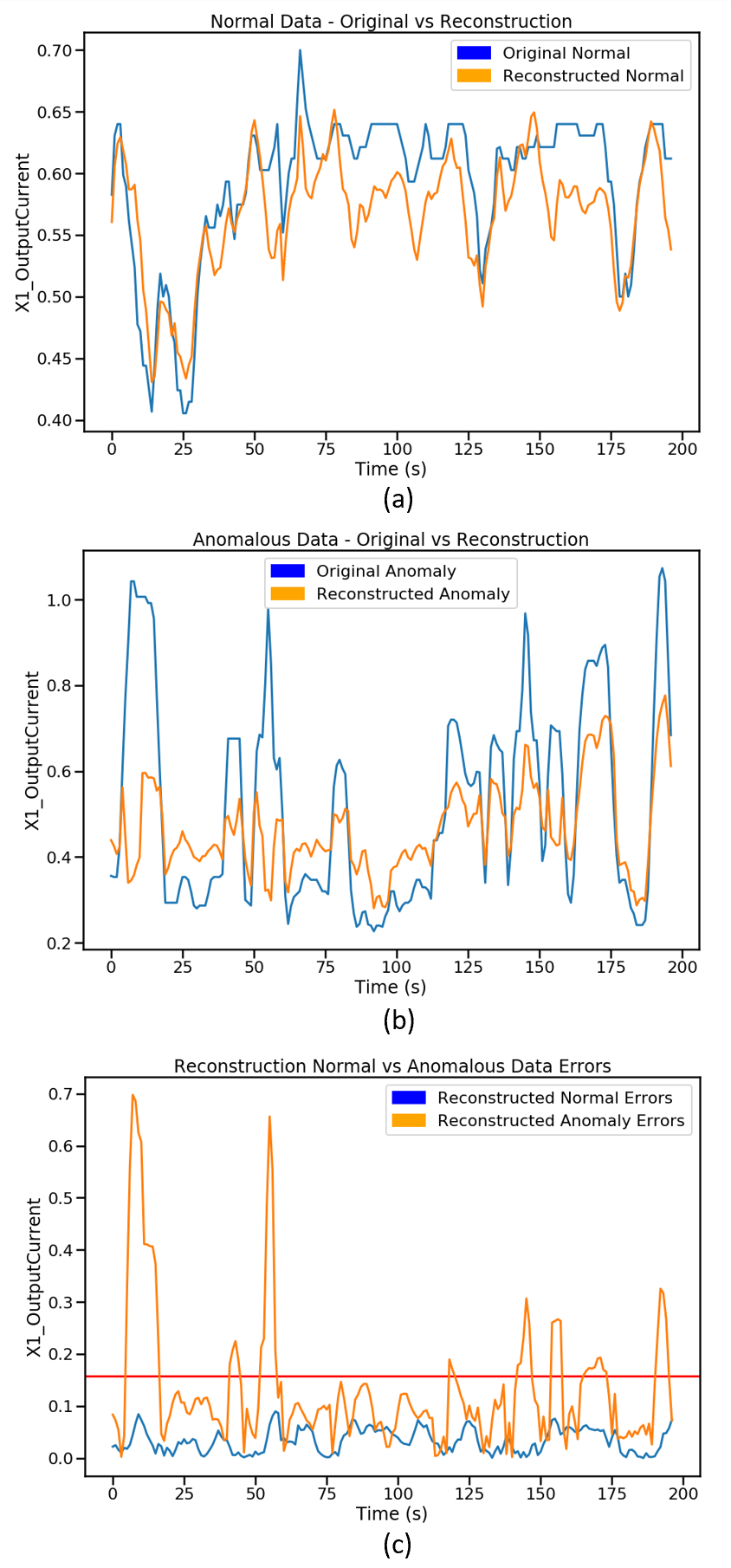}}
\caption{X1-OutputCurrent Time Series Measurement Analysis}
\label{fig:X1_OutputCurrent}
\end{figure}

X1-OutputCurrent was further analyzed in order to have a thorough understanding of ACLAE-DT's mechanism and results. Figure \ref{fig:X1_OutputCurrent} visualizes three charts within a specific time series cross-section: (a) X1-OutputCurrent original vs reconstructed normal data, (b) X1-OutputCurrent original vs reconstructed anomalous data, and (c) X1-OutputCurrent reconstructed normal vs anomalous data errors with the calculated dynamic threshold boundary in red. In chart (a), it can be observed that ACLAE-DT was able to reconstruct the original normal data well for most data points with a small margin of error. In chart (b), it can be observed that ACLAE-DT was not able to reconstruct the original anomalous data well, particularly for the reading peaks, as it never observed similar system statuses before. Finally, in chart (c), it can be realized that the reconstructed anomalous data errors crossed the threshold frequently, whereas the reconstructed normal data errors never crossed the threshold. It is important to note that many of the reconstructed anomalous errors were just shy of crossing the threshold, indicating that when the inter-correlation between X1-OutputCurrent and another time series was computed, the results were bound to cross the set threshold if the time series contained novel system behavior, contributing to X1-OutputCurrent's high anomalous correlation.


\subsection{Execution Time and Memory Requirements}
\begin{figure}[htbp]
\centerline{\includegraphics[width=9cm]{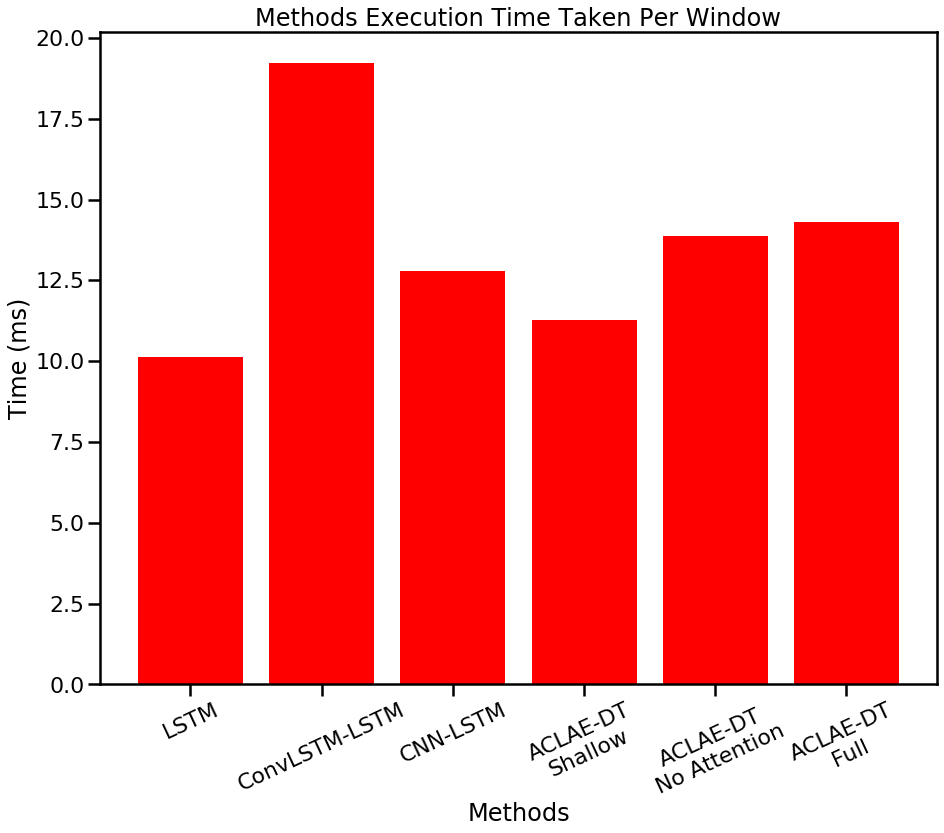}}
\caption{Execution Time Taken Per Window}
\label{fig:execution_time}
\end{figure}

\begin{figure}[htbp]
\centerline{\includegraphics[width=9cm]{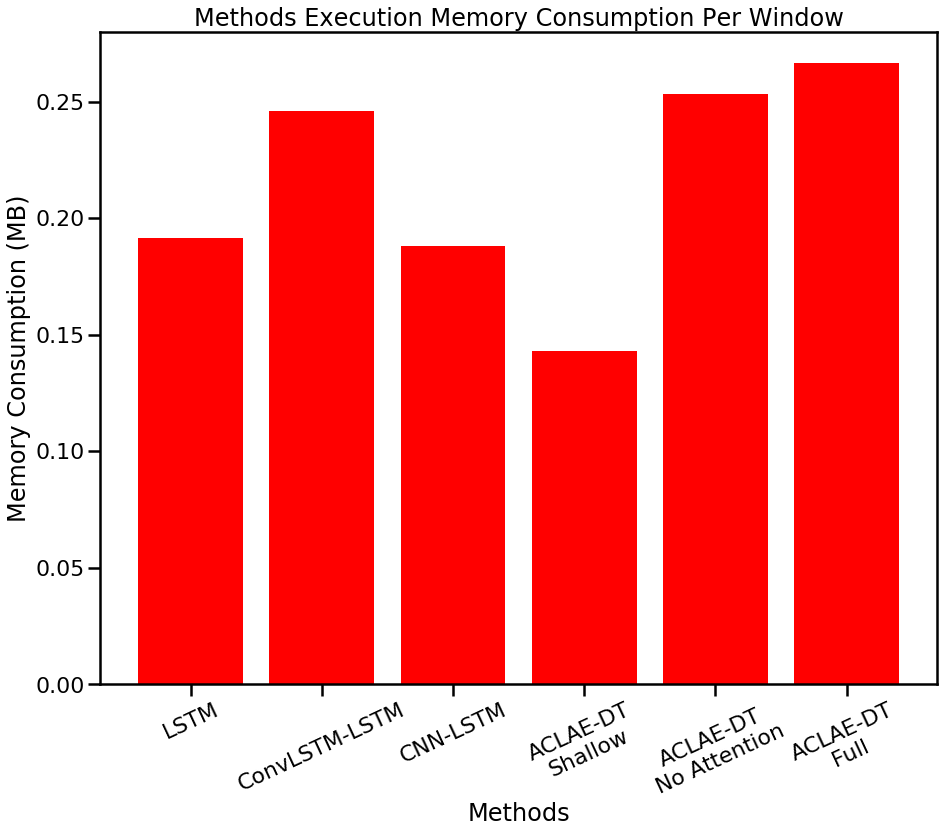}}
\caption{Execution Memory Consumption Per Window}
\label{fig:memory_consumption}
\end{figure}

To further evaluate ACLAE-DT against the baseline methods, the execution time and memory requirements of each method for a single window in Experiment 3 were calculated as visualized in Figures \ref{fig:execution_time} and \ref{fig:memory_consumption}, respectively. Ten experimental executions were conducted and the average results were used. SVM and ARIMA were not included in the comparison due to their poor anomaly detection performances, deeming both methods unsuitable for use in real-life SM processes. 

It can be observed from Figure \ref{fig:execution_time} that the LSTM autoencoder method took the shortest execution time per window with around 10 ms, followed by the ACLAE-DT shallow method with around 11.25 ms. The ACLAE-DT full method took around 14 ms, 40\% more time than the LSTM autoencoder method and 24.4\% more time than the ACLAE-DT shallow method. Moreover, it can be drawn from Figure \ref{fig:memory_consumption} that the ACLAE-DT shallow method required the least amount of memory during execution with around 0.14 MB. ACLAE-DT no-attention and ACLAE-DT full required the most amount of memory out of all methods, with memory consumptions of around 0.25 MB and 0.26 MB, respectively. From the drawn observations, it is evident that the tradeoff of using the full ACLAE-DT method with its superior anomaly detection performance, root cause detection identification, and short training time, is its greater execution time and memory consumption. If a method's execution time and memory consumption is of a higher importance than the method's anomaly detection performance and training time during real-life production, then the ACLAE-DT shallow method can be utilized as it performed very closely to the ACLAE-DT full method and required the second least amount of execution time and the least amount of execution memory out of all methods.

\section{Conclusion}\label{conclusion}
In this paper, a novel unsupervised attention-based deep ConvLSTM autoencoder with a dynamic thresholding mechanism framework, ACLAE-DT, was proposed to detect anomalies in a real-life manufacturing multivariate time series data set. The framework first normalized and enriched the raw time series with contextual information and sliding windows, before constructing feature images to capture system statuses across different time steps. The feature images were then input into an attention-based deep ConvLSTM autoencoder to be reconstructed, with an aim to minimize the reconstruction errors. The computed reconstruction errors were then subjected to a dynamic, nonparametric thresholding mechanism that utilized the mean and standard deviation of the normal reconstruction errors to compute a specific threshold for each time series pair, in order to detect and diagnose the anomalies.

Results demonstrated the effectiveness of ACLAE-DT, as it outperformed a classical approach, an ML approach, and three state-of-the-art DL approaches in detecting anomalous windows, while requiring less time to train than the latter approaches. Results further illustrated how ACLAE-DT was able to effectively diagnose the anomalies and locate the contributing features towards the anomalous windows. Moreover, the shallow variation of ACLAE-DT consumed the least amount of execution memory and the second least amount of execution time out the three state-of-the-art DL methods. All these results indicated the practicality and suitability of adopting ACLAE-DT in real life smart manufacturing processes. As a future extension to this work, ACLAE-DT can be applied to another public data set to benchmark its performance with the conventional anomaly detection algorithms, and a reduction in the execution time and memory consumption of the full ACLAE-DT model while maintaining the superior anomaly detection and anomaly root cause identification performances can be explored.


\section*{Acknowledgment}
This work is partially supported by the Natural Research Council (NRC) of the Government of Canada under Project AM-105-1.

\ifCLASSOPTIONcaptionsoff
  \newpage
\fi

\bibliographystyle{ieeetr}
\bibliography{references.bib}

\end{document}